\documentclass[sigconf,nonacm]{acmart}
\usepackage{gensymb}
\AtBeginDocument{%
  }

\begin{document}

\title{ECLIPSE: An Evolutionary Computation Library for Instrumentation Prototyping in Scientific Engineering}

\author{Max Foreback}
\email{foreba10@msu.edu}
\affiliation{%
  \institution{Michigan State University}
  \city{Lansing}
  \state{Michigan}
  \country{USA}
}

\author{Evan Imata}
\affiliation{%
  \institution{University of California, Berkeley}
  \city{Berkeley}
  \state{California}
  \country{USA}
}

\author{Vincent Ragusa}
\affiliation{%
  \institution{Michigan State University}
  \city{Lansing}
  \state{Michigan}
  \country{USA}
}

\author{Jacob Weiler}
\affiliation{%
  \institution{The Ohio State University}
  \city{Columbus}
  \state{Ohio}
  \country{USA}
}

\author{Jonathan Sy}
\affiliation{%
  \institution{University of California, Berkeley}
  \city{Berkeley}
  \state{California}
  \country{USA}
}

\author{Christina Shao}
\affiliation{%
  \institution{University of California, Berkeley}
  \city{Berkeley}
  \state{California}
  \country{USA}
}

\author{Joey Wagner}
\affiliation{%
  \institution{Michigan State University}
  \city{Lansing}
  \state{Michigan}
  \country{USA}
}

\author{Dylan Wells}
\affiliation{%
  \institution{The Ohio State University}
  \city{Columbus}
  \state{Ohio}
  \country{USA}
}

\author{Rick Marcusen}
\affiliation{%
  \institution{University of Colorado Boulder}
  \city{Boulder}
  \state{Colorado}
  \country{USA}
}

\author{Katherine G. Skocelas}
\affiliation{%
  \institution{Michigan State University}
  \city{Lansing}
  \state{Michigan}
  \country{USA}
}

\author{Aman Hafez}
\affiliation{%
  \institution{The Ohio State University}
  \city{Columbus}
  \state{Ohio}
  \country{USA}
}

\author{Amy Conolly}
\affiliation{%
  \institution{The Ohio State University}
  \city{Columbus}
  \state{Ohio}
  \country{USA}
}

\author{Kyle R. Helson}
\affiliation{%
  \institution{University of Maryland Baltimore County}
 \city{Catonsville}
 \state{Maryland}
 \country{USA}
}
\affiliation{%
  \institution{NASA Goddard Space Flight Center}
 \city{Greenbelt}
  \state{Maryland}
  \country{USA}
}

\author{Rajiv Ramnath}
\affiliation{%
  \institution{The Ohio State University}
  \city{Columbus}
  \state{Ohio}
  \country{USA}
}

\author{Wolfgang Banzhaf}
\affiliation{%
  \institution{Michigan State University}
  \city{Lansing}
  \state{Michigan}
  \country{USA}
}

\author{Charles Ofria}
\affiliation{%
  \institution{Michigan State University}
  \city{Lansing}
  \state{Michigan}
  \country{USA}
}

\author{Marcin Pilinski}
\affiliation{%
  \institution{University of Colorado Boulder}
  \city{Boulder}
  \state{Colorado}
  \country{USA}
}

\author{Bryan Reynolds}
\affiliation{%
  \institution{The Ohio State University}
  \city{Columbus}
  \state{Ohio}
  \country{USA}
}

\author{Anselmo C. Pontes}
\affiliation{%
  \institution{Autogenetics Research Lab}
  \state{Washington, DC}
  \country{USA}
}

\author{Emily Dolson}
\affiliation{%
  \institution{Michigan State University}
  \city{Lansing}
  \state{Michigan}
  \country{USA}
}

\author{Julie Rolla}
\affiliation{%
  \institution{Jet Propulsion Laboratory, California Institute of Technology}
  \city{Pasadena}
  \state{California}
  \country{USA}
}
\renewcommand{\shortauthors}{Foreback et al.}

\begin{abstract}
  
Designing scientific instrumentation often requires exploring large, highly constrained design spaces using computationally expensive physics simulations. These simulators pose substantial challenges for integrating evolutionary computation (EC) into scientific design workflows. EC typically requires numerous design evaluations, making the integration of slow, low-throughput simulators challenging, as they are optimized for accuracy and ease of use rather than throughput. We present ECLIPSE, an evolutionary computation framework built to interface directly with complex, domain-specific simulation tools while supporting flexible geometric and parametric representations of scientific hardware. ECLIPSE provides a modular architecture consisting of (1) Individuals, which encode hardware designs using domain-aware, physically constrained representations; (2) Evaluators, which prepare simulation inputs, invoke external simulators, and translate the simulator's outputs into fitness measures; and (3) Evolvers, which implement EC algorithms suitable for this domain. We evolve solutions for two novel space-science applications: 3D antennas optimized for directional sensitivity and spacecraft geometries optimized for drag reduction. Notably, we identify antennas with directional sensitivity roughly comparable to the expected sensitivity of two-antenna interferometric arrays, representing potential cost-savings. ECLIPSE enables interdisciplinary teams of physicists, engineers, and EC researchers to collaboratively explore designs for scientific hardware while leveraging existing domain-specific simulation software.
\end{abstract}




\maketitle

\section{Introduction}

Designing space science hardware is a complex, expensive, and highly iterative process. Traditional engineering workflows rely on expert-driven design followed by extensive simulation and refinement cycles. This process is time-consuming and can limit exploration of unconventional geometries. Evolutionary computation (EC) offers an appealing alternative: by automatically exploring large, high-dimensional design spaces, EC can discover high-performing and sometimes nonintuitive solutions that human designers may overlook \cite{onwubolu2013new}.

EC has been successfully applied to aerospace and electromagnetic design problems including evolved spacecraft components \cite{lohn2005evolved}, evolved antennas \cite{Machtay:2023Wm, rolla2023design, rolla2025designing, abarr2021payload}, and aerodynamic optimization \cite{sinpetru2022optimising}. However, applying evolutionary methods to new scientific hardware remains challenging. Most scientific simulation tools were not originally developed with evolutionary workflows in mind. They may be implemented in high-level languages, structured for serial execution, tightly coupled to specific data formats, or optimized for clarity rather than high-throughput evaluation. 

 The Nebulous collaboration is an interdisciplinary team of physicists, engineers, and computer scientists on the forefront of instrument and hardware optimization with a focus on space-science. Nebulous initially found success evolving 3D antennas from geometric primitives. Early work demonstrated that evolutionary methods could produce dipole-like antennas, and even design more sensitive antennas than those currently in use \cite{rolla2023design, rolla2025designing} when connected to a simulation that can predict performance on a science objective (in this case the number of observed ultra-high-energy neutrinos \cite{allison2015first}). As Nebulous expanded to tackle more complex problems, including more sophisticated antennas and spacecraft hardware, the demand for variable geometries with unforeseen constraints and assumptions was a common occurrence. As a result, the evolutionary workflow required significant maintenance to support the addition of new features. As integration of new components became increasingly time intensive we recognized that a single-purpose codebase was no longer sufficient. Inspired by software like the Modular Agent Based Evolver (MABE) \cite{bohm2017mabe}, we created ECLIPSE (Evolutionary Computation Library for
Instrumentation Prototyping in Scientific Engineering), a general framework for the evolutionary design of hardware in space-science domains.

A number of existing EC frameworks (e.g., DEAP \cite{fortin2012deap}, ECJ \cite{10.1145/3067695.3082467}, LEAP \cite{10.1145/3377929.3398147}, MABE \cite{bohm2017mabe}) provide flexible, well-tested implementations of standard EC algorithms and are widely used across research domains. However, these frameworks are designed as general-purpose toolkits and therefore leave the integration of complex scientific simulators largely to the user. In practice, this means that users must manually implement domain-aware representations, validity constraints, and communication with external scientific software. While this flexibility is valuable, it places a substantial engineering burden on domain scientists and can lead to ad hoc, problem-specific pipelines that are difficult to maintain or extend.

In contrast, ECLIPSE was designed specifically for scientific hardware design workflows. ECLIPSE is built around the assumption that simulators are externally maintained and owned by domain scientists rather than EC experts, and it aims to standardize the process of incorporating new simulators while reusing methods that mitigate the slow evaluation times and constrained parallelism typical of such tools.

ECLIPSE's interface standardizes how external simulators are invoked and how their outputs are processed. Classes for representing individual solutions are organized hierarchically to facilitate code reuse and support encoding problem-specific physical constraints. Classes for evolutionary algorithms, are engineered for evaluation regimes with extremely high computational cost and limited throughput. These conditions are typical of physics-based instrument modeling and Monte Carlo simulations of scientific phenomena. ECLIPSE provides the integration layer needed to couple evolutionary search with scientific modeling environments in a consistent, extensible, and domain-aware manner, enabling effective contribution from both EC experts and domain experts without requiring deep cross-disciplinary expertise. 

This paper provides an overview of ECLIPSE and its current capabilities. Section \ref{sec:framework} describes the modular architecture of the framework. Section \ref{sec:challanges} outlines the practical challenges encountered when integrating EC with domain-specific scientific simulation tools. Section \ref{sec:ongoing} presents two novel case studies from ongoing scientific applications the Nebulous collaboration is exploring using ECLIPSE. In the first, we evolve complex antennas on a multi-objective problem where we optimize both for a novel sensitivity goal and for simplicity. We find that our designs have the potential to achieve good performance at lower cost than the current solutions for this type of problem. In the second case study, we demonstrate a proof of concept that we can optimize CubeSat-scale spacecraft for very low Earth orbit (VLEO) drag characteristics. Section \ref{sec:conc} presents planned extensions for ECLIPSE and future work.

Upon request, the ECLIPSE framework may be accessed through collaboration with Nebulous, assuming all workflows can accommodate institutional restrictions.

\section{Challenges}
\label{sec:challanges}

Applying evolutionary computation to scientific hardware design introduces several practical challenges, many of which arise from integrating software and workflows developed across distinct research communities. Through our interdisciplinary work, we have identified three broad categories of obstacles: software interoperability, optimization and parallelization constraints, and the inherent computational cost of high-fidelity simulations.

\begin{figure*}
  \includegraphics[width=\textwidth]{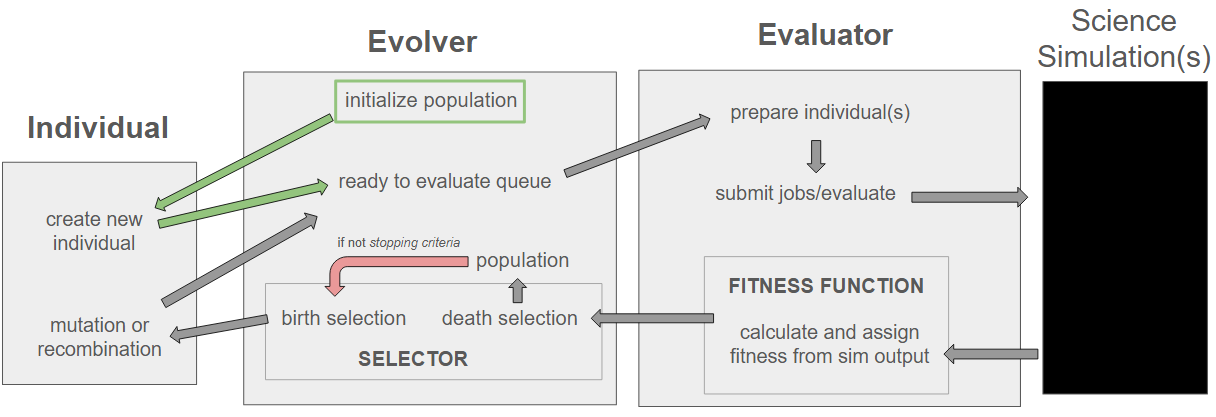}
  \caption{An overview of the ECLIPSE Framework. The algorithm begins and ends with the Evolver, which interfaces with an Evaluator module to get fitness scores, and an Individual module to receive new candidate solutions.}
  \label{wf}
  \Description{The workflow for the ECLIPSE algorithm.}
\end{figure*}

\subsection{Software Interoperability}
Simulation tools used in physics and aerospace research are typically developed with scientific correctness, reproducibility, and ease of model development in mind. As a consequence, they are often implemented in high-level languages like Python or MATLAB, which allow domain experts to rapidly prototype and validate physical models. These implementations are not optimized for the extremely high evaluation throughput needed for evolutionary computation. As a result, when these simulations are used as part of an evolutionary algorithm, their computational cost becomes one of the primary constraints.

Translating scientific simulators into lower-level, highly optimized languages is possible, but comes with significant verification, maintenance, and resource overhead. Translations often must be done by a developer without deep domain knowledge, and subtle errors that simple unit tests miss can easily be introduced. Additionally, ensuring correctness across independently maintained versions (e.g., a Python reference implementation maintained by domain experts and a C++ optimized version maintained by computer scientists) requires extensive validation infrastructure and ongoing synchronization efforts. Despite these challenges, the possibility of translating simulation software into lower-level languages should not be overlooked and may even be necessary if simulation times significantly limit the throughput of the EC algorithm. Future work aims to further explore the viability of simulation software translations and EC techniques, such as surrogate models \cite{jin2005comprehensive, dutta2020surrogate}, to reduce the computational burden put on the evolutionary pipeline by the Evaluators and science simulations.

ECLIPSE itself is written in Python for two primary reasons. First, the most computationally intense portions of the ECLIPSE pipeline are the third-party science simulations interfacing with Evaluators, which are imported as standalone software. Although writing Evolvers and Individuals in low-level languages such as C++ could theoretically provide higher performance, the benefit would be negligible due to the low relative cost of that portion of the framework. Second, low-level languages are not practical for collaborative development in our setting. Many contributors within Nebulous are domain scientists whose expertise lies in physics and instrument modeling rather than systems programming. Choosing a high-level language ensures that scientists can directly contribute modules, Evaluators, and test harnesses without forcing a steep tooling or language-learning burden. This choice significantly increases our collaboration's efficiency and reduces the barrier to entry for new scientific partners. 

\subsection{Parallelization and Workflow Constraints}
Most existing science simulation tools were not designed with evolutionary computation in mind. They often assume serial execution, maintain a persistent global state, or rely on I/O patterns that inhibit scaling to hundreds or thousands of parallel evaluations. In some cases, licensing or deployment restrictions on third-party simulation software also limits how it can be distributed across computing resources. These constraints shape the design of ECLIPSE and motivate the inclusion of evolutionary algorithms that remain effective even when evaluation budgets are severely limited.

\subsection{Inherent Computational Costs}
As previously discussed, the computational demands of evaluating candidate designs can be substantial. High-fidelity electromagnetic or aerodynamic simulations routinely require seconds to hours for a single evaluation and, in some extreme cases, even days. When combined with the aforementioned constraints on simulation software optimization and parallelization, evaluation cost becomes one of the dominant factors determining which evolutionary approaches are feasible. In the future, we plan to draw on the lessons learned in other evolutionary systems with expensive fitness functions to mitigate these challenges in ECLIPSE.\\

Collectively, these challenges highlight the difficulties of integrating evolutionary algorithms with domain-specific scientific simulations. They also illustrate why a framework such as ECLIPSE is needed. ECLIPSE enables scalable evolutionary optimization within the constraints of existing workflows and empowers contributions by domain experts without the need for in-depth knowledge of evolutionary systems and computing techniques.

\section{The ECLIPSE Framework}
\label{sec:framework}

The ECLIPSE Framework is divided into three types of modules. An \textit{Individual} module contains the genomic representation of the scientific hardware being evolved, as well as the representation's mutation and self-replication functions. An \textit{Evaluator} module manages communication between ECLIPSE and third-party science simulations written by domain experts. Evaluators are paired with \textit{fitness functions}, sub-modules that use the output of the simulation to calculate fitness scores. An \textit{Evolver} module manages a population of candidate solutions and handles birth and death selection through associated sub-modules called \textit{selectors}. The general workflow of ECLIPSE can be seen in Figure \ref{wf}.

\subsection{Individuals}

Individual modules define candidate designs in ECLIPSE and consist of their genetic representation and operators for mutation and recombination. Because ECLIPSE supports multiple types of hardware and relies on Evaluators built around domain-specific simulation tools, Individuals must capture the physical constraints that make a design valid for a particular problem. To accommodate this, ECLIPSE provides a flexible hierarchy of Individual types, allowing each problem domain to implement its own representation and mutation operators while still conforming to a shared interface used by Evolvers.

The most flexible class of Individual is the \textit{Shape Individual}, which builds 3D structures by combining primitive shapes (e.g., cuboids, cylinders) through a tree-like assembly process. \textit{Shape Individual} also provides general-purpose mutation operators that manipulate geometry (adding shapes, rotating shapes, etc.) while ensuring that the resulting design remains physically plausible.

Concrete subclasses refine this representation for use with specific Evaluators. For example, the \textit{Antenna Individual} extends \textit{Shape Individual} by introducing electrically conductive materials, voltage feed connections, and validation checks for electrical shorting. In contrast, the \textit{Spacecraft Individual} uses the same geometric foundation but enforces different constraints, namely fitting within specified bounding volumes and enforcing minimum cargo capacities. These representations share geometric mutation logic but implement their own domain-specific mutations, structural rules, and validation checks. Therefore, a given  Individual type can only be paired with an Evaluator capable of interpreting it: a spacecraft geometry cannot be evaluated by an electromagnetic solver, just as an antenna design cannot be passed to a drag simulator.

Not all representations derive from \textit{Shape Individual}. The \textit{Point Cloud Individual}, for instance, represents a spacecraft surface directly as a variable-length list of floating-point vertex coordinates. This surface respects maximum size constraints, and enforces a static internal cargo volume and static solar panels. Before evaluation verticies are translated into a watertight mesh using PyMeshLab \cite{pymeshlab}. This Individual type prioritizes flexibility and the tuning of fine-grained solutions that discrete representations would find harder to reach. All Individuals implement their own mutation operators, validity checks, and serialization logic while adhering to the standard interface expected by Evolvers and Evaluators.

By structuring Individuals in this way, ECLIPSE addresses a wide range of hardware design problems without imposing assumptions on what an Individual must be. It also allows experiments to incorporate domain knowledge as a starting point when appropriate. Each Individual type encapsulates the constraints and physical assumptions required by its corresponding Evaluator, while the shared interface allows the Evolver to treat all Individuals uniformly. This design makes it straightforward to add new representations as additional scientific domains are incorporated into the framework without altering the broader evolutionary workflow.

\subsection{Evaluators}

Evaluators are the central integration mechanism between ECLIPSE and third-party scientific simulation tools used to assess candidate designs. These simulators are standalone code written and maintained by domain experts. They accept a geometry or configuration, perform a high-fidelity physical simulation, and produce detailed output data. Because these tools were not originally designed with EC in mind, ECLIPSE does not modify or embed them directly. Instead, each Evaluator serves as a mediator between individuals in ECLIPSE and the external simulator that evaluates them.

The Evaluator module is responsible for preparing simulation inputs (e.g., generating 3D meshes or configuration files), invoking the external simulator, monitoring and validating execution, and parsing the resulting data products into a standardized form. Fitness functions, which remain modular and configurable, operate on the parsed output rather than on the raw simulation data. This separation allows domain scientists to retain their existing simulation code while enabling ECLIPSE to treat each simulator as a black-box module within the evolutionary workflow.

This architecture has several advantages. It isolates simulator-specific logic within a single component, simplifies the development of new problem domains, and maintains compatibility with complex legacy tools that cannot easily be rewritten or optimized for high-throughput evaluation. To maximize the number of candidate solutions that can be processed, all Evaluator code is asynchronous and returns control to Evolvers while awaiting the result of one or more simulations. ECLIPSE currently provides Evaluators for electromagnetic antenna simulation via XFdtd \cite{luebbers2006xfdtd} and spacecraft drag modeling via Vehicle Environment Coupling and TrajectOry Response (VECTOR) \cite{pilinski2011dynamic}. Additional Evaluators can be added as new scientific domains are incorporated into the framework, including surrogate models \cite{jin2005comprehensive}, which can serve as computationally efficient Evaluators when faster throughput is needed.

\subsection{Evolvers}

Evolvers are responsible for managing populations of candidate solutions and implementing the evolutionary algorithms used within ECLIPSE. At a high level, an Evolver defines the life cycle of an evolutionary run: how candidate solutions are selected, how new individuals enter the population, and how existing individuals are replaced. Unlike Evaluator modules, which integrate closely with domain-specific simulation tools, the design and implementation of Evolvers reside entirely within the domain of EC. Because Evolvers require no knowledge of the underlying physics, scientific objectives, or simulation details, they can be developed, optimized, and extended by experts in EC independently of the scientific components of a project. The high evaluation costs of most physics simulators mean that improvements in population management or search efficiency can directly translate to improved performance without changes to the simulation code.

ECLIPSE currently provides two Evolvers. The default is a steady-state genetic algorithm \cite{10.5555/93126.93169} incorporating an Age-Layered Population Structure (ALPS) \cite{10.1145/1143997.1144142, hornbySteadystateALPSRealvalued2009}. This Evolver ensures that new genetic material is continuously available to the population which can be useful in our computationally constrained runs, and aims to maintain diversity and mitigate premature convergence via injection from the ALPS regime. A simpler \textit{Hill-Climber Evolver} is also available for local search, fitness landscape exploration, and final optimization of evolved designs. Both Evolvers share the same modular interface, allowing them to be easily interchanged depending on the needs of an experiment.

Selection within ECLIPSE follows the same philosophy of modularity. An Evolver must be paired with a selection mechanism for choosing parents (birth selectors) and determining which individuals are replaced (death selectors). Many traditional selectors (e.g., tournament and roulette-based \cite{goldberg1991comparative}) are available, as well as a multiobjective selection scheme based on NSGA-II \cite{deb2002fast}, in which both parent and replacement decisions follow NSGA-II’s ranking and crowding calculations. 

The Evolver interface is designed to be lightweight and extensible. The difficult space-science problems the Nebulous collaboration is currently using ECLIPSE to solve are excellent applications for testing new EC techniques that make the most of a limited evaluation budget.

\section{Ongoing Work and Results}
\label{sec:ongoing}

ECLIPSE is actively being used across several scientific design applications that require high-fidelity simulation and complex geometric representations. As data for specific customers or missions is not publicly available, here we present case studies of two representative applications. In the first of these applications, antenna design, we seek to demonstrate that ECLIPSE can find a high-quality solution to a novel problem from a general domain that ECLIPSE has succeeded in before. In the second application, satellite design, we demonstrate a proof-of-concept that a novel point-cloud representation can be used to produce a reasonable solution in a problem domain that is fairly new to ECLIPSE. To demonstrate what a real-world user could reasonably expect from ECLIPSE, each of these case studies is a single run of the algorithm (physics simulations are generally too costly to enable running multiple replicates). Given this small sample size, we have taken great care to avoid introducing bias into our results. Both case studies here represent the first fully-functional run of ECLIPSE on their respective fitness functions. No problem-specific tuning of the algorithm or parameters has been applied, beyond those suggested by our initial intuitions. By avoiding additional runs of the algorithm, we have eliminated the risk of inadvertently cherry-picking results, and mirror the tight time constraints faced when ECLIPSE is utilized in practice. 

\subsection{Antenna Design}

Previous work by the Nebulous Collaboration has demonstrated that primitive-based representations can evolve dipole-like antennas for a science outcome (detection of ultra-high-energy neutrinos \cite{rolla2025designing}). Our current research extends this capability to more complex geometries by allowing the material associated with each primitive shape to mutate. If the material mutates to ``free space'', it is interpreted as air during simulation, enabling the evolution of passive reflector elements. If the material becomes ``feed'', it is replaced with a voltage feed connecting two pieces of conducting material, allowing for the evolution of antennas with multiple voltage feeds. 

These extensions significantly broaden the design space but also make the search problem more challenging, increasing the need for efficient evaluation strategies. To support this expanded parameter space, substantial performance improvements have been incorporated into ECLIPSE. Parallel Evaluator execution, refinements to both Evolver and Evaluator logic, and enhanced data integrity safeguards have collectively yielded an approximately 13-fold reduction in wall-clock time compared to experiments done prior to the integration of the ECLIPSE framework. These improvements enable substantially longer evolutionary runs, which are essential for discovering high-quality designs in complex spaces.

Here, we present a case study in which we test ECLIPSE on a novel antenna design problem.

\subsubsection{Antenna Design Results}
\label{ares}
\begin{figure}
  \includegraphics[width=0.5\textwidth]{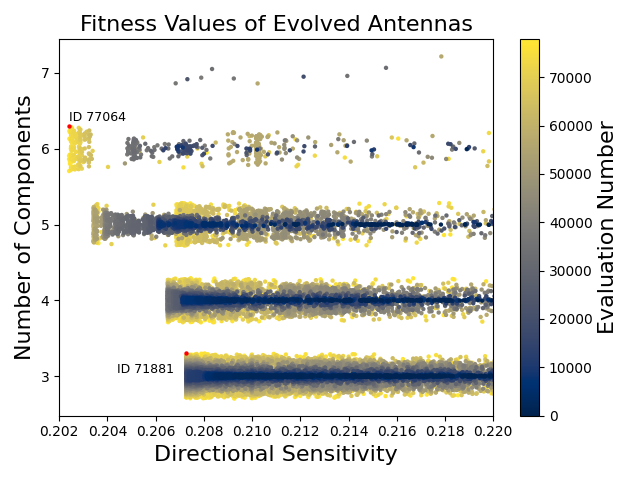}
  \caption{Directional sensitivity and number of components for individual solutions colored by time of evaluation. Both objectives are designed to be minimized. Fitness ranges from 0.2 to 0.8, but only elite solutions are shown here. Two individuals on the Pareto front are colored red and annotated by their individual IDs.}
  \label{pf}
  \Description{Fitness plot for evolved antennas.}
\end{figure}

\begin{figure*}
  \includegraphics[width=\textwidth]{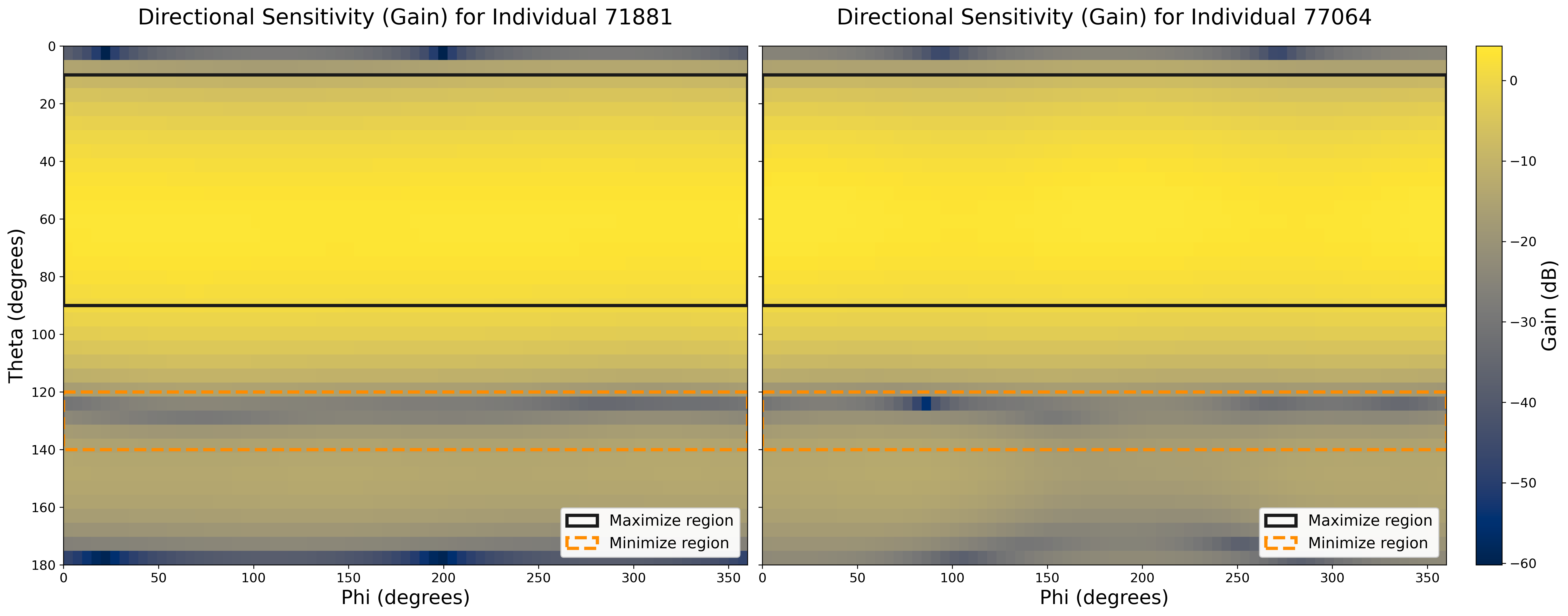}
  \caption{Evolved antenna gain patterns over observed regions for selected individuals. Higher gain for a direction results in a more sensitive antenna for that direction. Black regions are optimized for high gain, while orange regions are optimized for low gain. The values in these regions represent the input for equations \ref{max} and \ref{min} respectively.}
  \label{hm}
  \Description{Antenna sensitivity for selected antennas.}
\end{figure*}

\begin{figure}
  \includegraphics[width=.5\textwidth]{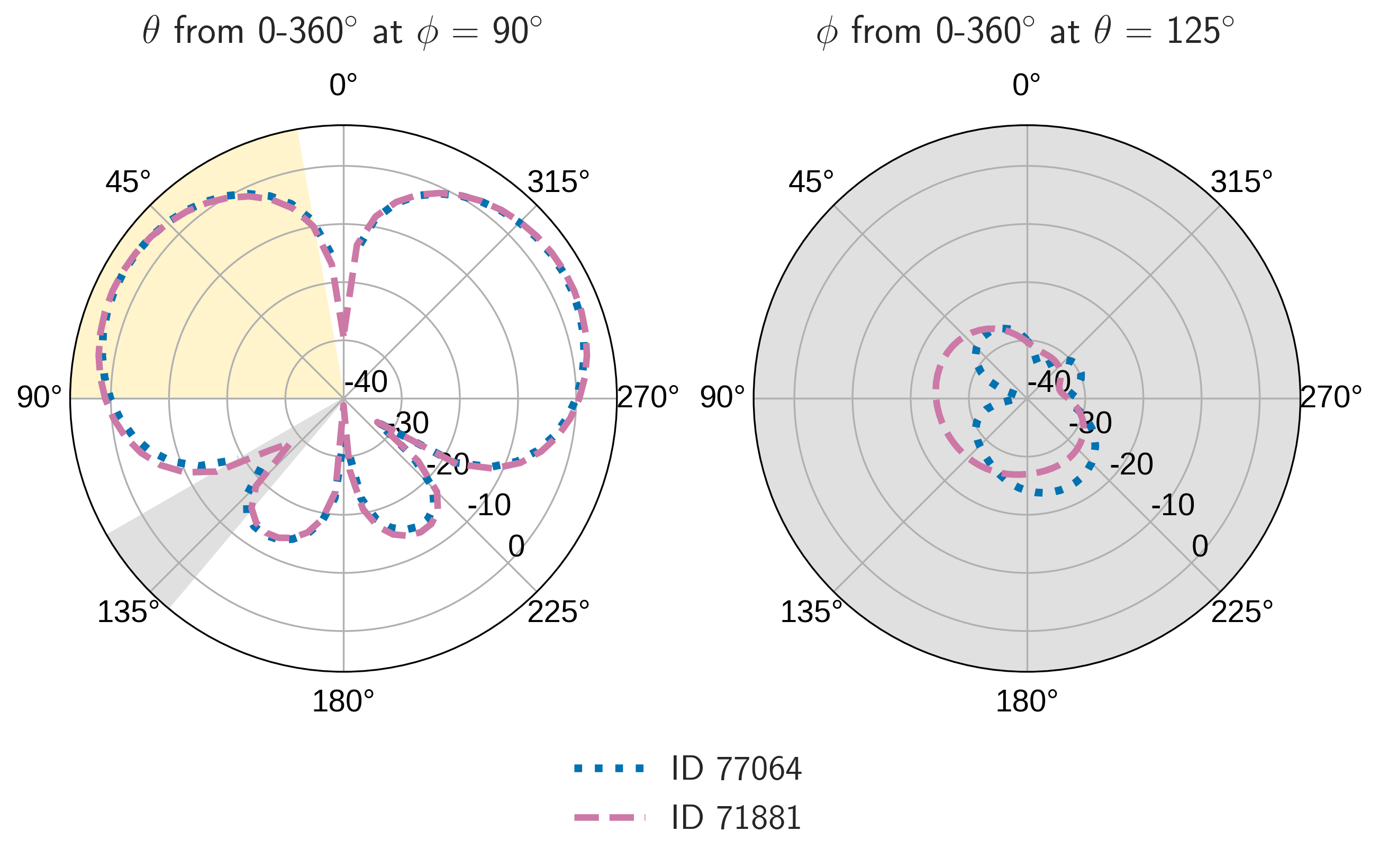}
  \caption{Gain plots of the evolved antenna patterns of individual 77064 (blue) and individual 71881 (pink). On the left, a vertical slice at $\phi = 90$ degrees shows the full range of $\theta$, highlighting the larger gain in the maximize region (yellow) and lower gain in the minimized region (gray) for both antennas. The right depicts a horizontal slice within the minimize region at $\theta = 125$, degrees where individual 77064 has the most extreme suppression of signal.}
  \label{slice}
  \Description{Antenna gain slices.}
\end{figure}

\begin{figure}
  \includegraphics[width=0.4\textwidth]{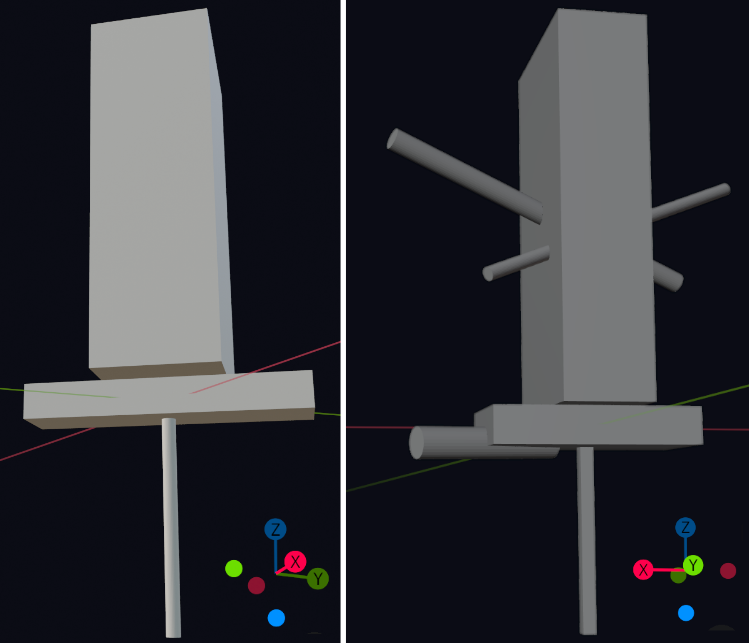}
  \caption{Evolved geometries of individual 71881 (left) and individual 77064 (right). In both of these individuals the horizontal cuboid connecting the top and bottom elements has the ``feed'' material type, and is replaced with a voltage feed on evaluation.}
  \label{indvs}
  \Description{Geometries of selected antennas.}
\end{figure}

\begin{table}
  \caption{Antenna run parameters. Parameters ending in ``percentage'' indicate the probability the corresponding mutation was chosen during a reproduction event.}
  \label{tennacfg}
  \begin{tabular}{ccl}
    \toprule
    Parameter& Value\\
    \midrule
    Number of layers & 3\\
    Layer size& 50\\
    Tournament size& 10\\
    Refresh interval& 2500\\
    Age progression& Linear\\
    Number of shapes& 3-20\\
    Allowed shapes& Cylinder, Cuboid\\
    Cuboid bounds (cm)& $1\leq l, w, h \leq40$\\
    Cylinder bounds (cm)& $0.5\leq r\leq3$, $8.75\leq h\leq40$,\\
    Rotation percentage& 10\%\\
    Dimension percentage& 40\%\\
    Regeneration percentage& 10\%\\
    Grow percentage& 20\%\\
    Prune percentage& 20\%\\
  \bottomrule
\end{tabular}
\end{table}

\begin{table}
  \caption{XFdtd run parameters.}
  \label{xf}
  \begin{tabular}{ccl}
    \toprule
    Parameter& Value\\
    \midrule
    Gain type & Realized\\
    Hollow shapes& True\\
    Hollow Thickness& 0.3 cm\\
    Frequency & 428 MHz\\
  \bottomrule
\end{tabular}
\end{table}

One of the key features of an antenna is its sensitivity in certain directions. As such, a fitness function is available to users that allows for the specification of directions in which to maximize and minimize the sensitivity of a given antenna. The directions of interest vary for each use case, but here we choose to maximize sensitivity between 10 degrees and 90 degrees in the theta direction, and minimize between 120 degrees and 140 degrees in theta, for all values of phi. Directional sensitivity is calculated as an average over all observed directions, rewarding high values in maximize regions and penalizing high values in minimize regions. Based on work done in \cite{10907045}, directional fitness is calculated as:
\begin{equation}
f = \max\left(\frac{1}{2}\left(\frac{1}{1 + \bar{G}_{\text{max}}} + \frac{\bar{G}_{\text{min}}}{1 + \bar{G}_{\text{min}}}\right), \epsilon\right)
\end{equation}

where:

\begin{equation}
\label{max}
\bar{G}_{\text{max}} = \frac{1}{|M|} \sum_{(\theta,\phi) \in M} 10^{g_\theta(\theta,\phi)/10}
\end{equation}

\begin{equation}
\label{min}
\bar{G}_{\text{min}} = \frac{1}{|m|} \sum_{(\theta,\phi) \in m} 10^{g_\theta(\theta,\phi)/10}
\end{equation}

$M$ is the set of direction points in maximize regions, $m$ is the set of direction points in minimize regions, $g_\theta(\theta,\phi)$ is the gain in dB at direction $(\theta, \phi)$, and $\epsilon = 10^{-12}$ as a small constant to prevent fitness from being zero. The parameters $\theta$ (theta) and $\phi$ (phi) are spherical coordinates that specify direction in 3D space. $\theta$ represents the elevation (zenith) angle from the vertical axis, ranging from 0° to 180°, while $\phi$ represents the azimuthal rotation angle in the horizontal plane, ranging from 0° to 360°. Together, ($\theta, \phi$) uniquely identify any direction vector pointing away from the antenna, allowing us to measure the antenna's gain in all directions.

In addition to sensitivity, the deployability of an antenna is an important constraint. As such, we also optimize for simple antennas, using the number of components as a proxy for simplicity. This run was performed with the steady state \textit{ALPS Evolver} and \textit{NSGA-II selector}, the \textit{Shape Individual} Individual type, and the \textit{XFdtd Evaluator}.

The constraints and challenges of a typical antenna optimization using ECLIPSE are replicated here. Runs which use the \textit{XFdtd Evaluator} are typically given a budget in wall clock time as opposed to elapsed generations, and the number of replicates are usually limited, as concurrent runs necessarily slow each other down through the use of shared API keys. Here, one replicate was performed and given two weeks of wall clock compute time on a supercomputer. Details of the run can be seen in Tables \ref{tennacfg} and \ref{xf}.

After two weeks of wall clock time, ECLIPSE completed just under 80,000 evaluations. Evaluated individuals and their corresponding fitness values (both to be minimized) and the time of their evaluation are shown in Figure \ref{pf}. We choose two designs on the Pareto front for further analysis, annotated in red in Figure \ref{pf}: individual 71881 with the fewest components, and individual 77064 with more components but the best overall directional sensitivity. 
Figure \ref{hm} presents the directional gain patterns for these two individuals, showing lower gain in the minimize region (orange dotted box) and higher gain in the maximize region (black outlined box).  

The better directional fitness of individual 77064 comes from its decreased sensitivity in the minimize region, most notable at around 100 degrees phi and 125 degrees theta (see Figure \ref{slice} for antenna response patterns of important regions). Figure \ref{indvs} shows the antenna geometries: individual 71881 is a vertical dipole, two vertical elements connected by a voltage feed, with a small cylinder on the bottom and a large cuboid on top. Individual 77064 has a similar base structure, a small vertical cuboid on the bottom electrically connected to a large cuboid on top, with the addition of crossing elements through the large top shape and an additional cylinder connected to the voltage feed. While reflectors were turned off for simplicity in these runs, these cylinders likely serve a similar purpose, dampening signal on the minimize region and yielding a better overall directional fitness score.

To understand the quality of this solution, it is useful to understand how directional responses are commonly achieved in space-based radio instruments.  The most common solution is achieved by combining the measurements from multiple simple antennas through interferometry or a phased array. Launching multiple antennas is very expensive and interferometry requires a great deal of precision in their placement, timing, and calibration. Both of our evolved antennas have similar gains throughout the maximize region of about 3.8 dB. As a rough comparison this level of gain is on the order of the maximum gain expected from a two-antenna phased dipole array. Thus for directional use cases, in theory, one of our evolved antennas could be launched into space in place of two antennas under the current state of the art paradigm, representing a substantial savings in cost and resources.

Based on design requirements an engineer could select the simpler antenna for deployability or prioritize the more complex antenna for the slight improvement within the minimize region. Final designs are often fine-tuned after an initial evolutionary run. Engineers typically manually align elements or remove unnecessary elements, and additional runs using the \textit{hill-climber Evolver} can be started with the evolved design as the starting point.

\subsection{Satellite Design}

\begin{table}
  \caption{VECTOR Parameters. Unlisted atmospheric composition values can be assumed to be 0 $m^{-3}$.}
  \label{vec}
  \begin{tabular}{ccl}
    \toprule
    Parameter& Value\\
    \midrule
    Temperature (K) & 1200.5\\
    Speed (m/s) & 7800.45\\
    Oxygen Composition (m$^{-3}$) & 1e11\\
    GSI Model & Fixed\\
    Energy Accommodation & 0.93\\
  \bottomrule
\end{tabular}
\end{table}

Satellites in very-low Earth orbit (VLEO) provide fast and affordable Earth imaging and sensing capabilities for applications like environmental monitoring and disaster response \cite{crisp2020benefits}. These satellites, however, experience substantial atmospheric drag, which strongly influences orbital lifetime, maneuverability, and mission reliability \cite{pilinski2011dynamic}. The topology of a satellite plays a major role in its drag profile, and past failures in this regime \cite{berger2023thermosphere} underscore the importance of accurately optimizing aerodynamic performance.

ECLIPSE is currently being used to evolve VLEO satellite topologies that minimize drag while satisfying mission-imposed structural constraints utilizing both primitive-based individuals and point cloud-based individuals. To reflect common VLEO mission configurations, the experiments use a 12U CubeSat as the baseline for size comparisons. Users can specify internal cargo volume requirements, overall bounding geometry, and atmospheric conditions.

As this domain is newer to ECLIPSE (and the representation we use here has not previously been tested), we present a proof-of-concept case study demonstrating ECLIPSE's ability to yield reasonable results on the short timelines often demanded by customers and missions.

\subsubsection{Satellite Design Results}
\begin{figure*}
  \includegraphics[width=\textwidth]{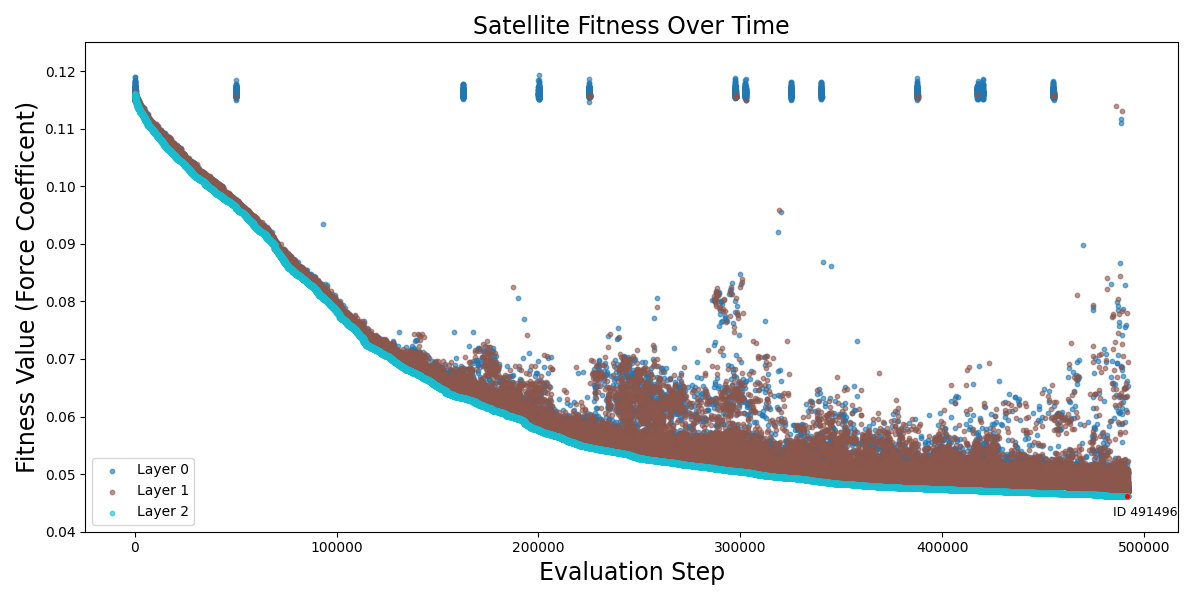}
  \caption{Evolved solutions through evolutionary time, colored by layer. The most elite individual in the final population is colored red. Hypothesized optimal performance (3U CubeSat) is shown as a dashed orange line.}
  \label{sobj}
  \Description{Evolved solutions through evolutionary time, colored by layer.}
\end{figure*}

\begin{figure}
  \includegraphics[width=0.45\textwidth]{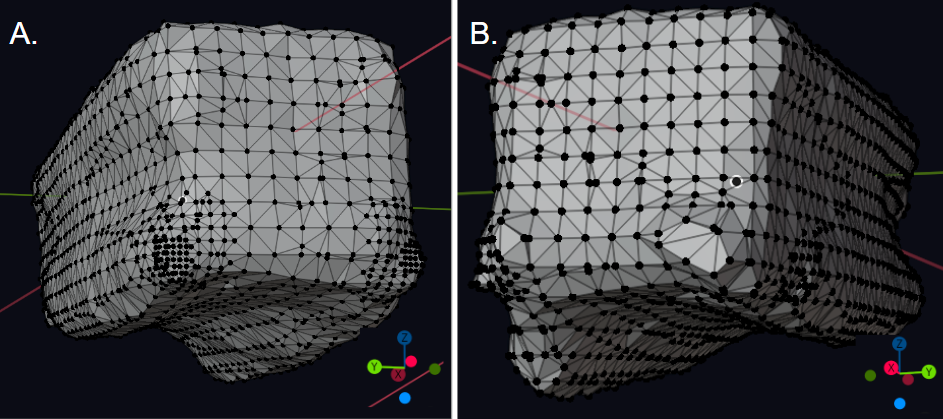}
  \caption{Best evolved CubeSat shown from the front (A) and back (B).}
  \label{sat}
  \Description{Geometries of evolved cubesat.}
\end{figure}

\begin{table}
  \caption{Satellite Run Parameters.}
  \label{satcfg}
  \begin{tabular}{ccl}
    \toprule
    Parameter& Value\\
    \midrule
    Number of layers & 3\\
    Layer size& 50\\
    Tournament size& 10\\
    Refresh interval& 2500\\
    Age progression& Linear\\
    Number of evaluations& 500,000\\
    Displacement range (m)& 0.005-0.02\\
    Minimum distance between vertices (m)& 0.001\\
    Initial mutations& 20\\
    Solar Panels& False\\
    Maximum Bounds (mm)& 200 x 200 x 300 (12U)\\
    Minimum Bounds (mm)& 100 x 100 x 300 (3U)\\
  \bottomrule
\end{tabular}
\end{table}

Here, we optimize a simple CubeSat for minimal drag as measured by the force coefficient as calculated by VECTOR. For details on calculating the force coefficient, see \cite{pilinski2011dynamic}. We use the \textit{Steady State ALPS Evolver} with simple tournament selection as the selector for this single objective problem. The VECTOR Evaluator was paired with the \textit{Point Cloud Individual} type, which allows evolution to fine tune individuals to a higher degree than \textit{Shape Individual}s. Run parameters can be seen in Table \ref{satcfg}. Evolution was allowed to run for 500,000 evaluations, with the fitness over time shown in Figure \ref{sobj}. While this initial proof-of-concept run did not achieve the theoretical optimal performance for our parameter settings (that of a 3U CubeSat, approximately 0.33), it got fairly close relative to its starting point, plateauing at a force coefficient of approximately 0.45. Given that these results reflect the first run of ECLIPSE on this problem (without any tuning of the algorithm or parameters based on prior problem-specific runs), we see these results as a promising demonstration of ECLIPSE's potential to be quickly applied to novel space science applications. 

The best individual found (shown in Figure \ref{sat}) has a curved underbelly and sides. This individual should be interpreted in the context of the specific environmental and modeling assumptions used for this run, found in Table \ref{vec}. VECTOR exposes a large number of parameters controlling atmospheric composition, flow conditions, and gas–surface interaction models, all of which directly influence the resulting force coefficients. In combination with ECLIPSE's flexible infrastructure, researchers can systematically permute these configurations while also allowing the evolutionary algorithm and individual representation to be adapted to the needs of a given problem. This set-up makes it straightforward to leverage the existing, highly configurable VECTOR simulation software within a coherent evolutionary pipeline, enabling researchers to flexibly swap evolutionary algorithms and individual representations to suit the needs of different orbital regimes or mission scenarios.

\section{Conclusion and Future Work}
\label{sec:conc}

ECLIPSE facilitates the integration of evolutionary computation into scientific workflows that rely on complex, domain-specific simulation pipelines. By separating representation, evaluation, and evolutionary logic, the framework enables interdisciplinary teams to explore unconventional hardware geometries while utilizing well-established scientific modeling tools. Although many challenges remain, the modular architecture of ECLIPSE provides a foundation upon which increasingly sophisticated optimizers and representations can be built. In this way, ECLIPSE seeks to accelerate existing engineering workflows and make evolutionary design a practical, routine component of scientific instrumentation development. Our results here demonstrate that ECLIPSE can 1) identify novel antenna designs with potential benefits over the current state of the art (in this case, likely cost savings), and 2) quickly be adapted to support new use cases and representations. It achieves both of these outcomes on the first try, without substantial problem-specific tuning or parameter sweeps ahead of time.

The Nebulous collaboration is currently developing several upgrades to ECLIPSE, focusing on the addition of new modules, and the improved efficiency and capability of existing modules. Plans are underway to allow for the evolution of an interferometric array of antennas which will vary not only the design of multiple antennas working together, but also their placement. Several upgrades to \textit{Shape Individual}s are planned as well, including new shapes, new ways of combining and connecting shapes, and new negative space mutations that allow for shapes to be hollowed out. Additional Individual types, including voxel-based representations \cite{baron1999voxel} and indirect representations (e.g., grammars \cite{o2002grammatical}, tree-based genetic programming \cite{koza1994genetic}), will be explored in the future. While the translation of scientific simulation software to low-level programming languages can be difficult, we also plan to implement a highly optimized C++ version of VECTOR \cite{pilinski2011dynamic} to increase the feasible number of evaluations when evolving spacecraft for drag. 

Additionally, techniques to mitigate the high cost of physics simulations are needed. In 2026, the Nebulous collaboration plans to integrate surrogate-assisted evolution and downsampling of expensive science simulations into ECLIPSE. Downsampling has been shown to efficiently make use of limited evaluations \cite{ferguson2020characterizing, 10.1145/3638530.3654208, 10.1145/3319619.3326900} and can be combined with surrogate modeling to drastically reduce computation time while preserving solution quality \cite{diaz2016review}.

These planned upgrades will build on the already strong foundation provided by the current version of ECLIPSE, facilitating a wide spectrum of new real world applications for evolutionary computation.

\begin{acks}
This research was supported by the National Science Foundation (NSF) through a Graduate Research Fellowship to MF (Award No. 2235783). Part of this work was carried out at the Jet Propulsion Laboratory, California Institute of Technology, under a contract with the National Aeronautics and Space Administration (80NM0018D0004). This work was also supported in part through computational resources and services provided by Remcom Inc. and the Ohio Supercomputer Center. The authors thank Olga Verkhoglyadova for advice on applications of this project. Any opinions, findings, conclusions, or recommendations expressed in this material are those of the author(s) and do not necessarily reflect the views of the NSF or affiliated institutions.
\end{acks}

\bibliographystyle{ACM-Reference-Format}
\bibliography{biblio}

@inproceedings{luebbers2006xfdtd,
  title={XFDTD and beyond-from classroom to corporation},
  author={Luebbers, Raymond},
  booktitle={2006 IEEE Antennas and Propagation Society International Symposium},
  pages={119--122},
  year={2006},
  organization={IEEE}
}

@phdthesis{pilinski2011dynamic,
  title={Dynamic gas-surface interaction modeling for satellite aerodynamic computations},
  author={Pilinski, Marcin D},
  year={2011},
  school={University of Colorado at Boulder}
}

@inproceedings{bohm2017mabe,
  title={MABE (Modular Agent Based Evolver): A framework for digital evolution research},
  author={Bohm, Clifford and G, Nitash C and Hintze, Arend},
  booktitle={Artificial Life Conference Proceedings},
  pages={76--83},
  year={2017},
  organization={MIT Press}
}

@techreport{rolla2025designing,
  author      = {Rolla, Julie and Reynolds, Bryan and Weiler, Jacob and Wells, Dylan and Foreback, Max and Connolly, Amy and Dolson, Emily and Ofria, Charles},
  title       = {{Designing Optimized Antennas for Science Applications Using Evolutionary Algorithms}},
  institution = {Jet Propulsion Laboratory},
  address     = {Pasadena, California},
  number      = {42-242},
  series      = {Interplanetary Network Progress Report: Volume 42-242},
  year        = {2025},
  month       = aug,
  pages       = {1-29},
  url         = {https://ipnpr.jpl.nasa.gov/progress_report/42-242/42-242A.pdf},
}

@techreport{rolla2023design,
  author      = {Rolla, Julie and Reynolds, Bryan and Weiler, Jacob and Connolly, Amy and Debolt, Ryan and Machtay, Alex and Sipe, Ben and Wells, Dylan},
  title       = {{Design of 3D Antenna Geometries Using Genetic Algorithms}},
  institution = {Jet Propulsion Laboratory},
  address     = {Pasadena, California},
  number      = {42-234},
  series      = {Interplanetary Network Progress Report: Volume 42-234},
  year        = {2023},
  month       = aug,
  pages       = {1-26},
  url         = {https://ipnpr.jpl.nasa.gov/progress_report/42-234/42-234A.pdf},
}

@article{deb2002fast,
  title={A fast and elitist multiobjective genetic algorithm: NSGA-II},
  author={Deb, Kalyanmoy and Pratap, Amrit and Agarwal, Sameer and Meyarivan, TAMT},
  journal={IEEE Transactions on Evolutionary Computation},
  volume={6},
  number={2},
  pages={182--197},
  year={2002},
  publisher={Ieee}
}

@inproceedings{10.1145/1143997.1144142,
author = {Hornby, Gregory S.},
title = {ALPS: the age-layered population structure for reducing the problem of premature convergence},
year = {2006},
isbn = {1595931864},
publisher = {Association for Computing Machinery},
address = {New York, NY, USA},
url = {https://doi.org/10.1145/1143997.1144142},
doi = {10.1145/1143997.1144142},
booktitle = {Proceedings of the 8th Annual Conference on Genetic and Evolutionary Computation},
pages = {815–822},
numpages = {8},
location = {Seattle, Washington, USA},
series = {GECCO '06}
}

@inproceedings{10.5555/93126.93169,
author = {Whitley, Darrell},
title = {The GENITOR algorithm and selection pressure: why rank-based allocation of reproductive trials is best},
year = {1989},
isbn = {1558600063},
publisher = {Morgan Kaufmann Publishers Inc.},
address = {San Francisco, CA, USA},
booktitle = {Proceedings of the Third International Conference on Genetic Algorithms},
pages = {116–121},
numpages = {6},
location = {George Mason University, USA}
}

@inproceedings{Machtay:2023Wm,
  author = "Machtay, Alexander Louis  and  Patton, Alex  and  Rolla, Julie Anne  and  Banzhaf, Wolfgang  and  Calderon, Dennis  and  Chen, Chi-Chih  and  Connolly, Amy  and  Debolt, Ryan  and  Fahimi, Ethan  and  King, Nick  and  Legersky, Maya  and  Melotti, Ezio  and  Reynolds, Bryan  and  Sipe, Ben  and  Staats, Kai  and  Stephens, Autumn  and  Tillman, Jack  and  Weiler, Jacob  and  Wells, Dylan  and  Wissel, Stephanie  and  Zinn, Audrey",
  title = "{Using Genetic Algorithms to Optimize Antenna Designs for Improved Sensitivity to Ultra-High Energy Neutrinos}",
  doi = "10.22323/1.444.1210",
  booktitle = "Proceedings of 38th International Cosmic Ray Conference {\textemdash} PoS(ICRC2023)",
  year = 2023,
  volume = "444",
  pages = "1210"
}

@book{sinpetru2022optimising,
  title={Optimising satellite geometries to minimise drag in Very Low Earth Orbits},
  author={Sinpetru, Luciana},
  year={2022},
  publisher={The University of Manchester (United Kingdom)}
}

@incollection{lohn2005evolved,
  title={An evolved antenna for deployment on nasa’s space technology 5 mission},
  author={Lohn, Jason D and Hornby, Gregory S and Linden, Derek S},
  booktitle={Genetic Programming Theory and Practice II},
  pages={301--315},
  year={2005},
  publisher={Springer}
}

@article{berger2023thermosphere,
  title={The thermosphere is a drag: The 2022 Starlink incident and the threat of geomagnetic storms to low earth orbit space operations},
  author={Berger, TE and Dominique, M and Lucas, G and Pilinski, M and Ray, V and Sewell, R and Sutton, EK and Thayer, JP and Thiemann, E},
  journal={Space Weather},
  volume={21},
  number={3},
  pages={e2022SW003330},
  year={2023},
  publisher={Wiley Online Library}
}

@book{onwubolu2013new,
  title={New optimization techniques in engineering},
  author={Onwubolu, Godfrey C and Babu, BV},
  volume={141},
  year={2013},
  publisher={Springer}
}

@incollection{goldberg1991comparative,
  title={A comparative analysis of selection schemes used in genetic algorithms},
  author={Goldberg, David E and Deb, Kalyanmoy},
  booktitle={Foundations of Genetic Algorithms},
  volume={1},
  pages={69--93},
  year={1991},
  publisher={Elsevier}
}

@inproceedings{hornbySteadystateALPSRealvalued2009,
  title = {Steady-State {{ALPS}} for {{Real-valued Problems}}},
  booktitle = {Proceedings of the 11th {{Annual Conference}} on {{Genetic}} and {{Evolutionary Computation}}},
  author = {Hornby, Gregory S.},
  date = {2009},
  series = {{{GECCO}} '09},
  pages = {795--802},
  publisher = {ACM},
  location = {New York, NY, USA},
  doi = {10.1145/1569901.1570011},
  url = {http://doi.acm.org/10.1145/1569901.1570011},
  urldate = {2014-08-12},
  isbn = {978-1-60558-325-9}
}

@article{crisp2020benefits,
  title={The benefits of very low earth orbit for earth observation missions},
  author={Crisp, Nicholas H and Roberts, Peter CE and Livadiotti, Sabrina and Oiko, Vitor Toshiyuki Abrao and Edmondson, Steve and Haigh, SJ and Huyton, Claire and Sinpetru, LA and Smith, KL and Worrall, SD and others},
  journal={Progress in Aerospace Sciences},
  volume={117},
  pages={100619},
  year={2020},
  publisher={Elsevier}
}

@article{jin2005comprehensive,
  title={A comprehensive survey of fitness approximation in evolutionary computation},
  author={Jin, Yaochu},
  journal={Soft Computing},
  volume={9},
  number={1},
  pages={3--12},
  year={2005},
  publisher={Springer}
}

@incollection{dutta2020surrogate,
  title={Surrogate model-driven evolutionary algorithms: Theory and applications},
  author={Dutta, Subhrajit and Gandomi, Amir H},
  booktitle={Evolution in Action: Past, Present and Future: A Festschrift in Honor of Erik D. Goodman},
  pages={435--451},
  year={2020},
  publisher={Springer}
}

@incollection{ferguson2020characterizing,
  title={Characterizing the effects of random subsampling on lexicase selection},
  author={Ferguson, Austin J and Hernandez, Jose Guadalupe and Junghans, Daniel and Lalejini, Alexander and Dolson, Emily and Ofria, Charles},
  booktitle={Genetic Programming Theory and Practice XVII},
  pages={1--23},
  year={2020},
  publisher={Springer}
}

@inproceedings{10.1145/3319619.3326900,
author = {Hernandez, Jose Guadalupe and Lalejini, Alexander and Dolson, Emily and Ofria, Charles},
title = {Random subsampling improves performance in lexicase selection},
year = {2019},
isbn = {9781450367486},
publisher = {Association for Computing Machinery},
address = {New York, NY, USA},
url = {https://doi.org/10.1145/3319619.3326900},
doi = {10.1145/3319619.3326900},
booktitle = {Proceedings of the Genetic and Evolutionary Computation Conference Companion},
pages = {2028–2031},
numpages = {4},
location = {Prague, Czech Republic},
series = {GECCO '19}
}

@inproceedings{10.1145/3638530.3654208,
author = {Lalejini, Alexander and Sanson, Marcos and Garbus, Jack and Moreno, Matthew Andres and Dolson, Emily},
title = {Runtime phylogenetic analysis enables extreme subsampling for test-based problems},
year = {2024},
isbn = {9798400704956},
publisher = {Association for Computing Machinery},
address = {New York, NY, USA},
url = {https://doi.org/10.1145/3638530.3654208},
doi = {10.1145/3638530.3654208},
booktitle = {Proceedings of the Genetic and Evolutionary Computation Conference Companion},
pages = {511–514},
numpages = {4},
location = {Melbourne, VIC, Australia},
series = {GECCO '24 Companion}
}

@article{diaz2016review,
  title={A review of surrogate assisted multiobjective evolutionary algorithms},
  author={D{\'\i}az-Manr{\'\i}quez, Alan and Toscano, Gregorio and Barron-Zambrano, Jose Hugo and Tello-Leal, Edgar},
  journal={Computational Intelligence and Neuroscience},
  volume={2016},
  number={1},
  pages={9420460},
  year={2016},
  publisher={Wiley Online Library}
}

@article{abarr2021payload,
  title={The payload for ultrahigh energy observations (PUEO): a white paper},
  author={Abarr, Q and Allison, P and Yebra, J Ammerman and Alvarez-Mu{\~n}iz, J and Beatty, JJ and Besson, DZ and Chen, P and Chen, Y and Xie, C and Clem, JM and others},
  journal={Journal of Instrumentation},
  volume={16},
  number={08},
  pages={P08035},
  year={2021},
  publisher={IOP Publishing}
}

@article{allison2015first,
  title={First constraints on the ultra-high energy neutrino flux from a prototype station of the Askaryan Radio Array},
  author={Allison, Patrick and Auffenberg, J and Bard, R and Beatty, JJ and Besson, DZ and Bora, C and Chen, C-C and Chen, Pnina and Connolly, A and Davies, JP and others},
  journal={Astroparticle Physics},
  volume={70},
  pages={62--80},
  year={2015},
  publisher={Elsevier}
}

@article{fortin2012deap,
  title={DEAP: Evolutionary algorithms made easy},
  author={Fortin, F{\'e}lix-Antoine and De Rainville, Fran{\c{c}}ois-Michel and Gardner, Marc-Andr{\'e} Gardner and Parizeau, Marc and Gagn{\'e}, Christian},
  journal={The Journal of Machine Learning Research},
  volume={13},
  number={1},
  pages={2171--2175},
  year={2012},
  publisher={JMLR. org}
}

@inproceedings{10.1145/3377929.3398147,
        Address = {New York, NY, USA},
        Author = {Coletti, Mark A. and Scott, Eric O. and Bassett, Jeffrey K.},
        Booktitle = {Proceedings of the 2020 Genetic and Evolutionary Computation Conference Companion},
        Doi = {10.1145/3377929.3398147},
        Isbn = {9781450371278},
        Location = {Canc\'{u}n, Mexico},
        Numpages = {9},
        Pages = {1571--1579},
        Publisher = {Association for Computing Machinery},
        Series = {GECCO '20},
        Title = {Library for Evolutionary Algorithms in Python (LEAP)},
        Url = {https://doi.org/10.1145/3377929.3398147},
        Year = {2020}}

@inproceedings{10.1145/3067695.3082467,
author = {Luke, Sean},
title = {ECJ then and now},
year = {2017},
isbn = {9781450349390},
publisher = {Association for Computing Machinery},
address = {New York, NY, USA},
url = {https://doi.org/10.1145/3067695.3082467},
doi = {10.1145/3067695.3082467},
booktitle = {Proceedings of the Genetic and Evolutionary Computation Conference Companion},
pages = {1223–1230},
numpages = {8},
location = {Berlin, Germany},
series = {GECCO '17}
}

@article{baron1999voxel,
  title={A voxel-based representation for evolutionary shape optimization},
  author={Baron, Peter and Fisher, Robert and Tuson, Andrew and Mill, Frank and Sherlock, Andrew},
  journal={Ai Edam},
  volume={13},
  number={3},
  pages={145--156},
  year={1999},
  publisher={Cambridge University Press}
}

@article{koza1994genetic,
  title={Genetic programming as a means for programming computers by natural selection},
  author={Koza, John R},
  journal={Statistics and Computing},
  volume={4},
  number={2},
  pages={87--112},
  year={1994},
  publisher={Springer}
}

@article{o2002grammatical,
  title={Grammatical evolution},
  author={O'Neill, Michael and Ryan, Conor},
  journal={IEEE Transactions on Evolutionary Computation},
  volume={5},
  number={4},
  pages={349--358},
  year={2002},
  publisher={IEEE}
}

@INPROCEEDINGS{10907045,
  author={Wells, Dylan and Rolla, Julie and Reynolds, Bryan and Connolly, Amy},
  booktitle={2025 United States National Committee of URSI National Radio Science Meeting (USNC-URSI NRSM)}, 
  title={Evolving Antennas for Directional Radio Sensitivity}, 
  year={2025},
  volume={},
  number={},
  pages={24-25},
  doi={10.23919/USNC-URSINRSM66067.2025.10907045}}

@software{pymeshlab,
  author       = {Alessandro Muntoni and Paolo Cignoni},
  title        = {{PyMeshLab}},
  month        = jan,
  year         = 2021,
  publisher    = {Zenodo},
  doi          = {10.5281/zenodo.4438750}
}
\end{document}